\documentclass{article}

\usepackage{arxiv}

\usepackage[utf8]{inputenc} 
\usepackage[T1]{fontenc}    
\usepackage{hyperref}       
\usepackage{url}            
\usepackage{booktabs}       
\usepackage{amsfonts}       
\usepackage{nicefrac}       
\usepackage{microtype}      
\usepackage{lipsum}
\usepackage{graphicx}

\usepackage{algorithm2e} 
\usepackage{algcompatible}
\RestyleAlgo{ruled}

\title{Neural Architecture Search by Learning a Hierarchical Search Space}

\author{
 Mehraveh Javan Roshtkhari \\
  Department of Systems Engineering\\
  École de technologie supérieure (ÉTS)\\
  Montreal, Canada \\
  \texttt{mehraveh.javan@gmail.com} \\
   \And
 Matthew Toews \\
  Department of Systems Engineering\\
  École de technologie supérieure (ÉTS)\\
  Montreal, Canada  \\
  \texttt{matthew.toews@etsmtl.net} \\
  \And
 Marco Pedersoli \\
  Department of Systems Engineering\\
  École de technologie supérieure (ÉTS)\\
  Montreal, Canada  \\
  \texttt{marco.pedersoli@etsmtl.ca} \\
}

\begin{document}
\maketitle
\begin{abstract}
Monte-Carlo Tree Search (MCTS) is a powerful tool for many non-differentiable search related problems such as adversarial games.
However, the performance of such approach highly depends on the order of the nodes that are considered at each branching of the tree. If the first branches cannot distinguish between promising and deceiving configurations for the final task, 
the efficiency of the search is exponentially reduced. 
In Neural Architecture Search (NAS), as 
only the final architecture matters, the visiting order of the branching can be optimized to improve learning. 
In this paper, we study the application of MCTS to NAS for image classification. We analyze several sampling methods and branching alternatives for MCTS and propose to learn the branching by hierarchical clustering of architectures based on their similarity. The similarity is measured by the pairwise distance of output vectors of architectures.
Extensive experiments on two challenging benchmarks on CIFAR10 and ImageNet show that MCTS, if provided with a good branching hierarchy, can yield promising solutions more efficiently than other approaches for NAS problems.

\end{abstract}

\section{Introduction}
\label{intro}
Neural Architecture Search (NAS) aims to automate neural architecture design and has shown great success in the recent years~\cite{zoph2016neural, liu2018progressive,real2019regularized,ren2021comprehensive}, surpassing manually designed Convolutional Neural Networks (CNN) in deep learning ~\cite{liu2018darts,liu2019auto,guo2020single}.
NAS aims at yielding the best architecture within a given search space with a lower computational budget than a brute-force approach, based on training all possible architectures independently.

One prominent solution is one-shot methods based on weight sharing~\cite{pham2018efficient} in which multiple architectures share all or part of their weights. 
These methods circumvent the need for individual training of each architecture by training a single "supernet" that contains all possible operation/architectural choices. Each architecture are evaluated by using weights inherited from that supernet.
When architectures are compatible, weight-sharing allows to recycle training iterations~\cite{cha2022supernet, pham2018efficient,bender2018understanding}; however, for vastly different architectures, it may lead to interference, i.e. weights beneficial to one harm others and vice-versa~\cite{roshtkhari2023balanced}. 

To reduce the detrimental effect of interference in weight sharing, prior work has explored either multiple specialized models that can focus on different parts of the search space~\cite{su2021k,zhao2021few,hu2022generalizing,roshtkhari2023balanced}; or importance sampling~\cite{liu2018darts,ye2022b,xu2019pc} in which the probability of an architecture performing well is estimated during training. In the latter, promising architectures are sampled more frequently during the course of training, gradually reducing the possible interference~\cite{liu2018darts,you2020greedynas} as the training is guided towards the best architectures. Unlike uniform sampling~\cite{guo2020single,roshtkhari2023balanced}, the challenge of using importance sampling is to robustly estimate and identify superior architectures as early as possible in the training cycle. This would allow the model to focus on these architectures and to minimize wasting training resources on unpromising architectures. This requires fast and reliable estimation of the probability distribution of architectures in as few training iterations as possible. 

\begin{figure*}
    \centering
    \begin{tabular}{c c c} \includegraphics[width=0.18\linewidth]{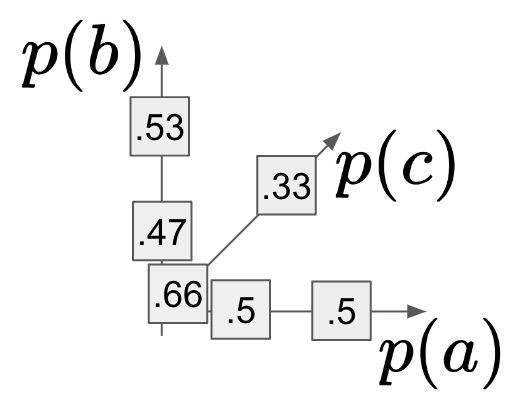} \hspace{1.5cm} & \includegraphics[width=0.18\linewidth]{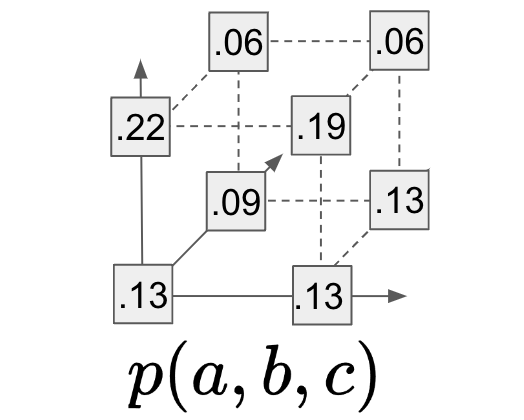} & \hspace{1.5cm} \includegraphics[width=0.18\linewidth]{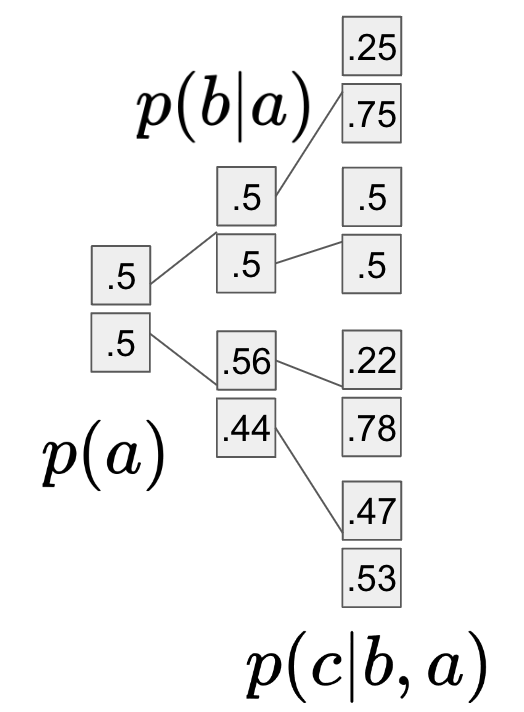}\\(independent)\hspace{1.5cm} & (joint) & \hspace{1.5cm}(conditional)\\
    \end{tabular}
    
    \caption{\textbf{Probability factorization of 8 architectures.} We show different ways to approximate the discrete probability distribution of architectures for a toy example of search space with N=3 nodes (a,b,c in the figure) each one with O=2 possible operations for a total of $2^3$ architectures. (left) Assuming the nodes independent (as in DARTS \cite{liu2018darts}) allows the model to estimate only $N \times O$ probabilities. (center) Considering the joint probabilities would require to estimate $O^N$ different probabilities (as in Boltzmann sampling).
    (right) The joint probability can be factorized into the product of conditional probabilities (in a hierarchy such as in MCTS). This does not reduce the probabilities to estimate, but allows a more efficient exploration of the search space.}
    \label{fig:prob}
\end{figure*}

A neural network can be viewed as a graph, composed of nodes, which define the architecture, connected by edges. These nodes offer a choice of operations, which are the processes applied to the data (e.g. convolution, fully connected, etc.). In importance sampling, a common assumption for more efficient estimation of architecture probability is "node independence", i.e. considering nodes as statistically independent variables. For instance, in a neural network, the choice of the operation for the second layer would not depend on the choice of the operation in the first layer. This simplifies architecture probability estimation to a product of nodes' probabilities (see Fig. \ref{fig:prob} (independent)). This reduces the scale of the problem to learning individual node probabilities. While the widely used differentiable NAS method (DARTS~\cite{liu2018darts} and followup works~\cite{xu2019pc,li2020sgas,ye2022b}) rely on this assumption, overlooking the joint contribution of nodes to architecture performance can lead to a poor node selection for the final architecture~\cite{ma2023inter,zhang2024dependency}.

In order to remove the node independence assumption, joint probabilities of all configurations should be estimated (see Fig. \ref{fig:prob} (joint)). This requires estimating each architecture's probability independently. In mainstream single-path one-shot (SPOS) methods~\cite{guo2020single,chu2021fairnas, li2020random,stamoulis2019single}, at each iteration, one architecture from the supernet is sampled, trained and estimated and that estimation is used to update the probability distribution. With node independence, at each iteration several associated node probabilities are updated; while for joint probability only the probability of sampled architecture is be updated. 
Thus, the full update of the estimation costs proportionally to the number of architectures, making it unscalable to large search spaces. 
This inefficient estimation slows down the entire training as the importance sampling will not be able to focus on the promising architectures. 


To explore more wisely, a compelling option is to factorize the joint probability into conditionals, such as a tree structure for Monte-Carlo Tree Search (MCTS) (Fig. \ref{fig:prob} (conditional)). While the number of probabilities to estimate remains the same as joint probability, it offers several key advantages. A well-designed  hierarchical search space enables more efficient tree traversal by reducing the unnecessary exploration of unpromising branches. Prioritizing promising branches can lead to faster convergence, improved solution quality and better scalability. 
However, standard predefined hierarchies do not guarantee a more efficient exploration, as they are defined by construction and without taking into account the semantic similarity of the corresponding architectures.

While in many MCTS problems the search hierarchy is defined by the sequentiality of the problem (e.g the moves in chess), for NAS there is no constraint in the order of exploring architectures.
\cite{zhao2021multi} and \cite{wang2021sample} leverage this by using the classification accuracy of the nodes for partitioning the search space into "good" and "bad" nodes to reduce the unnecessary exploration. This approach works well when running the search with a fixed, already learned, recognition model (i.e. the CNN weights). 
In fact, \cite{zhao2021multi} uses MCTS for searching the best performing model on a supernet pre-trained with uniform sampling, while \cite{wang2021sample} perform MCTS for NAS using the already trained models provided by NAS-Bench-201~\cite{dong2020bench}.

In this work, we tackle the more challenging problem of learning the recognition model and the tree partitioning jointly. In this setting, at the beginning of the optimization, the recognition model has poor performance, and using its accuracy as proxy to partition the search space would be ineffective. 
Instead, we propose to estimate the distance between architectures in an unsupervised manner, using a partially trained recognition model, and without relying on the model accuracy.
The output vector of each architecture, sampled from this supernet, is used to calculate a pair-wise distance matrix of architectures. We propose to use this matrix for hierarchical clustering and generating tree partitions. The resulting hierarchical clustering implicitly enforces that the early nodes of tree to be semantically related, without directly factoring their performance. This approach enhances learning and consequently accelerates the search process.

The main contributions of our work are the following:
\begin{itemize}
    \item We present a new understanding of classical choices of models and strategies for NAS based on the sampling approach and the estimation of the underlying probability of a given architecture. We show that overly restrictive assumptions (e.g. node independence) enables faster training, but converges to suboptimal solutions. In contrast, adopting more realistic assumptions, combined with additional regularization can lead to better solutions. 
    \item We propose an efficient method to sample from search tree by learning to construct a good hierarchy that avoids low-performing architectures. 
    To build this hierarchy, we evaluate several approaches and show that the
    most promising one is obtained based on pairwise distances between architectures, derived from a supernet pre-trained with uniform sampling.
    \item We empirically validate our findings on two NAS benchmarks on CIFAR10 dataset and mobilenet ImageNet search space. Our results show that the proposed approach is very general and can discover promising architectures within a limited computational budget.  
\end{itemize}

\section{Related Work}

\textbf{One-shot methods.}
One-shot methods~\cite{pham2018efficient,bender2018understanding} has become very popular in NAS \cite{liu2018darts,guo2020single,su2021prioritized} due to their efficiency and flexibility.
 Generally, the training of supernet and searching for best architecture can be decoupled \cite{guo2020single,wang2021sample} or performed simultaneously \cite{liu2018darts}. In the former, the search can be performed by various methods, such as random search \cite{bender2018understanding,li2020random}, evolutionary algorithms \cite{guo2020single} or MCTS \cite{wang2021sample} and the supernet is static during this phase. The latter alternates between training the supernet and updating the reward to guide the search, such as updating architecture weights in differentiable methods \cite{liu2018darts}, controller in RL \cite{pham2018efficient} or probability distribution in MCTS \cite{su2021prioritized}. 
The quality of supernet as a proxy for architecture evaluation has been the subject of scrutiny in recent years, with various results in different settings \cite{yu2019evaluating,wang2021rethinking,zela2019understanding,termritthikun2021eeea,zhang2024boosting}. A proposed solution is to explicitly reduce the weight sharing among architectures by non-hierarchical factorization of the search space \cite{zhao2021few, roshtkhari2023balanced, su2021k}. However, in general these methods are computationally more expensive as they require training additional models. Tree based approaches can be viewed as a form of hierarchical factorization of the search space that reduces weight sharing.



\textbf{Node independence.}
Early NAS methods using reinforcement learning \cite{zoph2016neural,pham2018efficient}, or evolution \cite{real2019regularized, sun2020automatically, sun2019evolving} do not treat nodes as independent, but they were computationally expensive. More efficient and widely adopted NAS methods are differentiable methods (based on DARTS \cite{liu2018darts}) that use back-propagation to learn both node weights (probabilities) and supernet weights. However, one of their known issues is that the learned weights for the nodes fail to accurately reflect their contribution to the ground truth performance and ranking \cite{wang2021rethinking,yu2019evaluating}. 
While several studies has recognized and attempted to improve DARTS \cite{chen2021progressive,ye2022b,xu2019pc}, few have directly explored the contribution of node independence assumption to this problem \cite{ ma2023inter,xiao2022shapley}.

Shapley-NAS \cite{xiao2022shapley} highlights the underlying relationship between nodes by showing that the joint contribution of node pairs often differs from the accumulation of their separate contributions, due to  potential collaboration/competition. They propose to reweigh the learned architecture weights using Shapley value. However, estimating the Shapely value can be costly as it requires training the supernet multiple times. ITNAS \cite{ma2023inter} proposes to explicitly model the relationship between nodes by introducing a transition matrix and an attention vector that denotes the node probability translation to successor nodes. The matrix and vector are optimized in a bi-level framework alongside node probabilities. However, the application of this method is limited to only cell level (inner layer operations) and extension to a more general macro search space is not straightforward (for details about macro and cell-based search spaces see Appendix \ref{app:related}).
While these works try to incorporate node dependencies into differentiable NAS, an alternative approach is to directly learn either joint or conditional probabilities of the sampled architectures.

\textbf{Monte-Carlo Tree Search.} 
\label{sec:MCTS}
MCTS with Upper Confidence bound applied to Trees (UCT) \cite{auer2002finite} has been used previously for NAS \cite{negrinho2017deeparchitect,wistuba2017finding}. AlphaX \cite{wang2019alphax} used a 
surrogate network to predict the performance of sample architectures, and MCTS to guide the search. 
TNAS \cite{qian2022meets} aims to improve the exploration of the search space by using a bi-level tree search that traverses layers and operations iteratively. However, the binary tree that factorizes the operations is manually designed.
LaMOO \cite{zhao2021multi} and LaNAS \cite{wang2021sample}  aim to tackle the problem of finding the best architecture 
and assume that deep leaning model is given either from a trained supernet \cite{wang2021sample} or using precomputed benchmarks \cite{zhao2021multi}.
In our work, we instead aim at training the deep learning model and finding the corresponding optimal architecture with MCTS in the same optimization, which makes the problem more challenging.


The closest work that jointly performs the model training and architecture search with MCTS is \cite{su2021prioritized}
that propose to construct a tree branched along operations. During the training, a hierarchical sampling is used for node selection, updating the supernet weights and the reward (training loss). Node statistics are then used to update a relaxed UCT probability distribution. However, the tree design is manual and a regularization method is required to compensate for insufficient visits of nodes. 




\section{Training by Sampling Architectures}
\label{sec-method}
Our training is based on a SPOS \cite{guo2020single} in which, given a neural model $f$ (e.g. a CNN) for each mini-batch of training data $\mathcal{X}$ and corresponding annotations $\mathcal{Y}$, a different architecture $a$ from the search space $\mathcal{S}$ is sampled and back-propagated with the following loss:
\begin{equation}
    \mathcal{L}(f_a(\mathcal{X},w),\mathcal{Y}) = \sum_{(x,y) \in (\mathcal{X,Y})} l(f_a({x},w),{y}),
\end{equation}
where $l$ is the sample loss (for instance cross-entropy) and $w$ is the network weights.  
Training speed and model performance can vary significantly depending on how $a$ is sampled. To prevent overfitting on training data in importance sampling methods, we use validation accuracy as the reward to estimate the probability distribution; and use on-line estimation on mini-batches to accelerate the process. In the following sub-sections, we present some of the most common sampling techniques, from uniform sampling to our proposed approach.

\textbf{Uniform sampling.}
\label{method:uniform}
The simplest and the original approach of SPOS \cite{guo2020single}, in which an architecture is sampled uniformly: $a \sim \mathcal{U}(|\mathcal{S}|)$, with $|\mathcal{S}|$ denoting the cardinality of the search space. Despite its simplicity, this sampling method is unbiased (not privileging specific architectures) and given enough training, all architectures will have the same importance. This method also requires no additional information storage during the training, and in principle can accommodate any $|\mathcal{S}|$, even very large ones. 
In practice however, the equal importance of architectures can present two possible challenges: i) With strong weight sharing (i.e. most of the model weights are shared among all configurations), the same weight would have to adapt to very different architectures,leading to destructive interference and therefore low performance (see \cite{roshtkhari2023balanced}). ii) If weight sharing is minimal, each architecture is almost independent and the training time would increase proportionally to $|\mathcal{S}|$. While uniform sampling may be combined with search space partitioning in a trade-off \cite{roshtkhari2023balanced}, it requires training multiple models, demanding higher memory consumption and computational cost.
A different direction is to find ways to prioritize the sampling of the more promising architectures. 

\textbf{Importance sampling with independent probabilities.}
\label{method-independent}
The simplest way to estimate the importance of each operation is assuming each node $a_i$ as independent. Thus, the probability of an architecture $a$ is approximated as $p(a)=p(a_1)p(a_2)...p(a_t)$. This simplifying assumption improves sampling efficiency by factorizing the number of probabilities to estimate. However, the quality of the solution is compromised, as it disregards the joint influence of nodes on the performance \cite{ma2023inter}. 

\textbf{Importance sampling with joint probabilities: Boltzmann sampling.}
\label{method-Boltzmann}
In Boltzmann sampling, architecture $a$ is sampled from a Boltzmann distribution with probability $p(a)
\propto \exp(\frac{\epsilon_a}{T}) $, where $\epsilon_a$ is estimated rewards (here accuracy) of $a$, and $T$ is the temperature. Sampling is performed with an annealing temperature, starting from a high value (almost uniform), such that the initial phase of the training is unbiased, to a low value (almost categorical) such that the training focuses on high performing architectures. 
While more efficient than uniform sampling, estimating $\epsilon_a$ remains time consuming, particularly for large search spaces and it is difficult to balance exploration/exploitation trade-off in Boltzmann exploration \cite{cesa2017boltzmann}. 

\begin{figure*}[t]
    \centering
    \begin{tabular}{c c c} \includegraphics[width=0.18\linewidth]{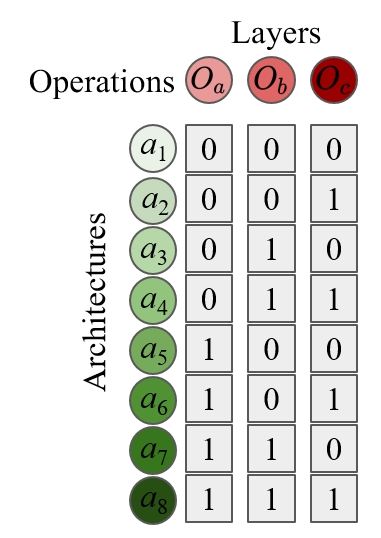} \hspace{1.5cm} & \includegraphics[width=0.18\linewidth]{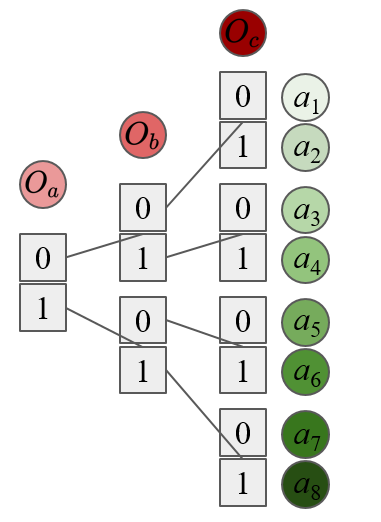} & \hspace{1.5cm} \includegraphics[width=0.18\linewidth]{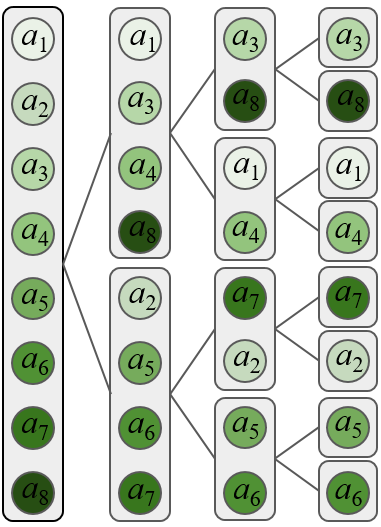}\\(a) Search space\hspace{1.5cm} & (b) Default tree & \hspace{1.5cm}(c) Learned tree\\
    \end{tabular}
    
  \caption{\textbf{Comparison of the standard tree structure and our learned structure on a 3 binary operations search space.} 
  (a) The search space consists of architectures with 3 binary operations ($o_a,o_b,o_c$) which leads to 8 architectures ($a_1,a_2,...,a_8$). (b) The default tree structure uses the order of operations (e.g. layers) to build the tree, however this is not optimal. (c) Our learned tree structure uses a tree that is generated by an agglomerative clustering on the model outputs. 
  }
 \label{fig:tree_design}
\end{figure*}

\textbf{Sampling with conditional probabilities: Tree Search.}
Instead of a flat vector of probabilities, we consider a tree of conditional probabilities: $p(a) = p(a_t|a_{(t'\le t-1)})p(a_{t-1}|a_{(t' \le t-2)})...p(a_1)$. Each $a_t$ represents a level of the tree that partitions the set of possible architectures into disjoint subsets. The commonly used structure of the tree (Fig.\ref{fig:tree_design}~(b)) is defined by factorizing the model architectures layer by layer \cite{su2021prioritized}, starting from the first to the last one. 
Assuming for simplicity a symmetric binary tree, the first level would split the configurations into two disjoint groups. This process of recursive partitioning continues at each subsequent level. With uniform sampling, at each iteration the nodes in level $t$ are sampled with probability $(1/2)^t$.

Thus, probability estimates for early nodes tend to be sufficiently accurate because of high sampling rate. In contrast, for Boltzmann, the sampling rate is ${1}/{|\mathcal{S}|}$, which can be extremely small for a large search space. 
However, if initial nodes maintain a near-uniform probability distribution (not sufficiently discriminative), the estimation of the posterior nodes would suffer from low sampling rate.
A possible solution is the regularization proposed by \cite{su2021prioritized}, in which at each update of a specific node, all other equivalent nodes (nodes at the same level with the same operation) are updated similarly using an exponential moving average. This mechanism of multiple simultaneous updates allows for faster probability estimation and more efficient exploration.

While seemingly an adequately solution, this regularization comes with limitations: i) It assumes homogeneous tree structure at each level, i.e. all nodes at a given level have similar structure (identical children), which limits the approach to specific of search spaces. For instance, this approach would not be suitable for search spaces where the operations in a node are conditioned to the choice of operation at the previous node. ii) Reusing the same probabilities for equivalent nodes implies treating nodes independently. In this case, the node independence assumption is enforced in a soft way by a regularization coefficient. Thus, the method attempts to find a compromise between full independence and conditional dependence, but it remains unclear if this trade-off is optimal. 

\section{Our Approach}
\label{method-our}
Our approach tackles the low sampling rate issue in posterior nodes of MCTS from another perspective. We aim to find an ordering of nodes for the tree such that, especially for the initial levels of the tree, the probabilities of a sub-node are imbalanced. 
This enables the search to concentrate on a reduced set of architectures as the unpromising branches would be estimated early and sampled less frequently.
Considering the example in Fig. \ref{fig:prob}~(conditional), instead of building the tree from node $a$, we could start from node $c$. In this case, the imbalanced probability of $c$ enables to focus effectively on a good branch, thus making the sampling much more efficient.
This is possible because, unlike other problems in which the ordering of nodes is defined as part of the problem, NAS only cares about the final architecture and not the order used to reach it.
Therefore, in this work we propose to learn an improved ordering of nodes to sample in a MCTS, instead of using a predefined ordering.

In this section, we present a different approach to build a tree of architectures.
As shown in Fig.\ref{fig:tree_design}~(c), our tree is built based on hierarchical clustering of architectures. Each node of the tree represents a cluster of architectures, going from the root that contains all architectures in a single cluster to the leaves that each contains a single architecture. Through this approach, we release the tree from dependence on the binary operations and allow any possible hierarchical grouping of architectures. 

\begin{table*}
    \caption{\textbf{Accuracy and ranking on the Pooling benchmark on CIFAR10.} We report the found architecture (represented with number of layers per feature map sizes), best and average of 3 training accuracy and ranking and search time for different methods.} 
    \label{Toy-comparison}
    \vskip 0.1in
    \centering
    \begin{tabular}{l c c c c c c}
       Method  & Arch. &  Best Acc.  &  Avg.  Acc. &  Best Rank &  Avg. Rank & Search Time\\
       \hline
         Default Arch. & [4,3,3] & 90.52 & - & 15 & - & -  \\
         \hline
         Uniform & [4,3,3]  & 90.52 & $90.40 \pm 0.08$ & 15 & 17 & 1.5\\        
         MCTS &[4,4,2]  & 90.85 & $90.57 \pm 0.21$ & 12 & 15.3  & 2\\
         Boltzmann & [3,5,2] & 90.88 & $90.51 \pm 0.12$ & 11 & 15.3 & 3 \\
      Independent & [3,5,2] & 90.88 & $90.86 \pm 0.01$ & 11 & 11.7 & 2 \\
         Mixtures \cite{roshtkhari2023balanced} & [5,3,2] & 91.55 & $91.36 \pm  0.27 $ & 4 & 5 &  6 \\
         MCTS + Reg.\cite{su2021prioritized} & [6,1,3]  & 91.78  & $91.42 \pm  0.11 $ & 3 & 3.6  & 2\\
         MCTS + Learned (ours) & [6,2,2] & 91.83 & 91.72 $\pm 0.12$ & 2 & 3 & 2\\
         \hline
         Best & [7,1,2] & 92.01 & - & 1 & - &  -\\
    \end{tabular}
\end{table*}

In this work, we build a hierarchy that takes into account the semantic similarity between different architectures.
A representation of architectures that is independent of the quality of supernet, while adequately summarizes the relevant information for the task, can be used to determine the placement of architectures in search space. A clustering algorithm can then be used to build the hierarchy based on the distances between architectures. We note that while the ultimate goal of NAS is to find the architecture with the highest accuracy, the supernet accuracy by itself is an inadequate representation of architecture for this purpose. In contrast, the output vector is more suitable, as distances between architectures would have semantic meaning in the class space
(See sec.\ref{ablation} for ablation studies on tree design).

First, a supernet is pre-trained with uniform sampling for a specified number of iterations. Then, we sample architectures from the supernet and perform a forward pass with a mini-batch of validation data and record the output vector. Next, The pairwise distances of architectures are calculated and the resulting distance matrix is used for hierarchical agglomerative clustering \cite{murtagh2014ward} to construct a binary tree. 
We argue that this method allows us to effectively cluster architectures with similar overall functionality, even if they might differ in their structure in term of their operations. 
For more details about the construction of the tree, see algorithm 1 in Appendix \ref{app:exp_details}.

\textbf{Search and training.}
We use a modified MCTS for both supernet training and architecture search. Similar to \cite{su2021prioritized,wang2021sample}, the tree is fully pre-expanded and thus expansion and roll-out stages of traditional MCTS are skipped. Similar to \cite{su2021prioritized}, we use Boltzmann sampling for the selection stage. The Boltzmann distribution allows sampling proportional to the probability of the reward function, producing better exploration \cite{painter2024monte}, which is fundamental for good training of the model. For a node in the tree $a_i$, we perform importance sampling with a Boltzmann sampling relative to the node:

\begin{equation}
\label{eq:uct}
    p(a_i) = \frac{\exp(R(a_i)/ T)}{\sum_j \exp(R(a_j)/ T)} ,
\end{equation}

where $R$ is our reward function and $T$ is temperature, determining the sharpness of distribution and the normalization sum on $j$ is on the sibling nodes of $i$.
Training consists of sampling each level of the tree from the root to the leaf, followed by a gradient update of the recognition model $w$ with the sampled architecture $a$ on a mini-batch of training data $\mathcal{X}_{tr}$ and an update of the reward function for the explored nodes, from the leaf to the root to the tree based on the obtained accuracy of the architecture on a mini-batch of the validation set $\mathcal{X}_{val}$.
To balance exploration/exploitation, the Upper Confidence bound applied to Trees (UCT) \cite{kocsis2006bandit} is used as reward for sampling. Considering a node in tree $a_i$, our reward is defined as:

\begin{equation}
\label{eq:rew}
R(a_i) = C(a_i) + \lambda \sqrt{{log(|parent(a_i)|})/{|a_i|}} ,
\end{equation}

in which the second term is for exploration. We use function $|a_i|$ to show number of times node $a_i$ is visited, with the constant $\lambda$ controlling the exploration/exploitation trade-off. $parent(a_i)$ indicates the parent node of $a_i$. We define $C$ as:

\begin{equation}
    C(a_i) = \beta ~ C(a_i) + (1-\beta) ~ Acc(f_a(\mathcal{X}_{val},w)),
\end{equation}

which is a smoothed version of the validation accuracy of architecture $a$, with smoothing factor $\beta$, to account for the noisy on-line estimation on mini-batches. 
For further details about the training algorithm see algorithm 1 in Appendix \ref{app:exp_details}.
To search for the final architecture after training, we sample $k$ architectures without exploration ($\lambda = 0$) and rank them based on their performance on validation dataset, selecting the best as the final architecture. 

\section{Experiments}
\label{sec-exp}
We evaluate our method on CIFAR10 dataset \cite{krizhevsky2009learning} using 
two macro search space benchmarks, 
 and ImageNet \cite{ILSVRC15} with MobielNetV2-like \cite{sandler2018mobilenetv2} search space . 
We compare our proposed method with several various sampling methods discussed in sec. \ref{sec-method}. 
In all experiments, we use SPOS method for sampling and training the supernet. 
For MCTS methods, we start by uniform sampling of the architectures, and after a warm-up period, we use the recorded statistics to calculate UCT (eq. \ref{eq:rew}) and sample using eq. \ref{eq:uct}. 
For MCTS default tree, used by \cite{su2021prioritized}, each layer of CNN is considered as a level of tree, with operations providing the branching. We compare with this method with and without soft node independence assumption (regularization). For further experimental details see Appendix \ref{app:exp_details}.

\subsection{Pooling Search Space}
\label{Toy-Search}
To fully investigate our proposed method, we use Pooling search space, a small yet challenging CIFAR10 benchmark consisting of 36 Resnet20-like \cite{he2016deep} architectures. 
The only architecture parameter that is optimized is feature map sizes at each layer (or equivalently the placement of downsamplings operations). Therefore, at each layer the choices are whether to perform downsampling or not (pooling or Identity operation).  The main challenge of this search space lies in full weight sharing among architectures that contributes to the inadequacy of several common search methods\cite{roshtkhari2023balanced}.
Here, we represent architectures with number of layers per feature map sizes (e.g. [4,3,3] meaning 4/3/3 layers in high/middle/low resolution). Our method achieves better results in comparison within similar or shorter search time (Table \ref{Toy-comparison}). While for MCTS (default tree design), the regularization proposed in \cite{su2021prioritized} appears to help, our method obtains better performance without needing regularization.

\begin{figure}
  \centering
  \includegraphics[width=0.50\linewidth] {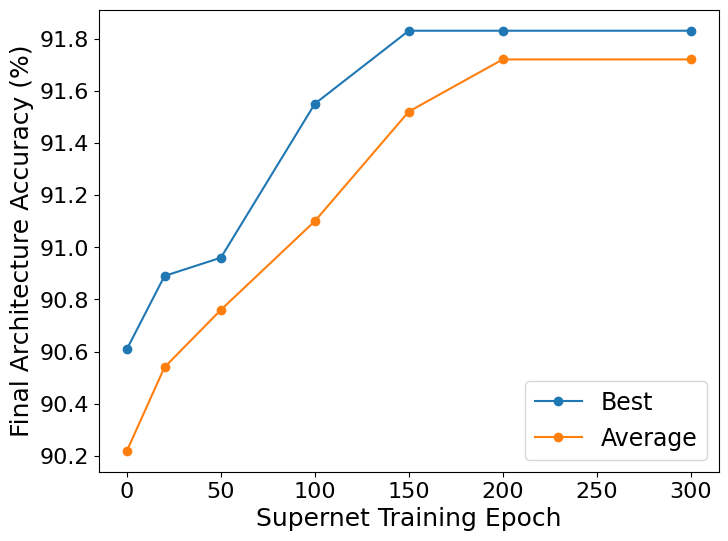}
  \includegraphics[width=0.48\linewidth]
  {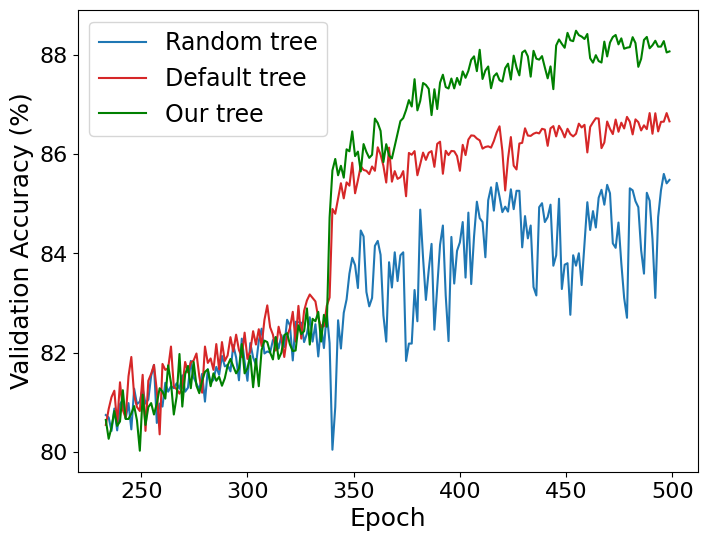}
  \caption{(left) \textbf{Training epochs for estimating the similarity matrix.} We show the final performance of our MCTS in which the tree structure is learned with a model uniformly trained for a given number of epochs. For best results at least 200 epochs are needed; (right) \textbf{Accuracy over epochs for several training strategies.} After the warm-up phase, our approach is constantly better than default tree or MCTS with a randomly selected tree.}
  \label{supernet-iter}
\end{figure}

\subsection{Ablation Studies}
\label{ablation}
We conduct several ablations to assess the convergence and performance of our method and explore alternative ways to design the hierarchy in Pooling search space. 

\textbf{Tree partitioning with accuracy.} 
We investigate the importance of using the output vector and clustering, we build a tree using the accuracy obtained from a pre-trained supernet. We recursively partition search space into "good" (top 50\% of the partition) and "bad" regions (bottom 50\%). Performing MCTS on this tree, we observed diminished results compared to our method, with best and average accuracy of $90.85\%$ and $90.01\%$ respectively.

\textbf{Clustering distance measures.}
Using CNN's output to calculate pairwise distances, the difference between two distributions can be calculated by several distance measures. In table \ref{distance-metric}, we investigate three common distnace measures: L2 distance, KL divergence, and cross-entropy.

\begin{table}[tb]
    \caption{\textbf{Distance measures for the similarity matrix.} We compare the final performance of our MCTS with learned tree structure which is built with an agglomerative clustering using a similarity matrix between network outputs with different distance measures.}
    \vskip 0.1in
    \label{distance-metric}
    \centering
    \begin{tabular}{c c c c}
       Distance Measure & \begin{tabular}{@{}c@{}}Best \\ Arch. \end{tabular} & \begin{tabular}{@{}c@{}}Best \\ Acc. \end{tabular} & \begin{tabular}{@{}c@{}}Avg. Acc. \end{tabular}\\
       \hline
        cross-entropy & [5,3,2] & 91.55 & $91.20 \pm 0.23$\\
        L2 & [6,2,2] & 91.83 & $91.52 \pm 0.16$\\
         KL & [6,2,2] & 91.83 & $91.72 \pm 0.12$ \\ 
    \end{tabular}
\end{table}
\textbf{Supernet training and similarity matrix.}
During supernet training, the average accuracy typically increases. In our experiments, we used supernet at convergence. However, since the criteria for our proposed clustering is similarity and not the exact accuracy, supernet needs to be trained only long enough to capture that similarity. To investigate the impact of supernet training duration, we evaluated our method with various training iterations (see Fig.\ref{supernet-iter}(left)). Notably, even without any training, our method is able to find a better architecture than the default, and with only 1/3 of full training, we can find better architecture compared to \cite{roshtkhari2023balanced}. This observation suggests a potential trade-off between supernet pertaining budget and search budget.

\begin{table}
    \caption{\textbf{Comparing various zero cost branching methods.} We consider one-hot encoding of operations per layer or the categorical vector representation. The similarity matrix is calculated using L2 distance. We also consider an exponential weighting scheme to increase the influence of earlier layers on distance. }
    \vskip 0.1in
    \label{Toy-alt-branch}
    \centering
    \begin{tabular}{c c c c c}
       Encoding & Weighted & \begin{tabular}{@{}c@{}}Best \\ Arch. \end{tabular} & \begin{tabular}{@{}c@{}}Best \\ Acc. \end{tabular} & Avg. Acc. \\
       \hline
         Vector &           & [2,5,3] & 90.89 & $90.63 \pm 0.68$ \\
         Vector &  \checkmark & [3,4,3] & 90.92 & $90.81 \pm 0.15$ \\
         \hline
         One-hot &          & [5,2,3] & 90.96 & $90.60 \pm 0.22$ \\
         One-hot & \checkmark & [5,1,4] & 91.05 & $90.78 \pm 0.21$  \\
    \end{tabular}
\end{table}

\begin{table*}
    \caption{\textbf{Accuracy and ranking on NAS-Bench-Macro.} We compare our method and several approaches in terms of best, average accuracy and ranking. The architectures are represented with operation index per layer. 
    }
    \vskip 0.1in
    \label{Bench-Macro-comparison}
    \centering
    \begin{tabular}{l c c c c c c}
       Sampling  & Arch. &  Best Acc.  &  Avg. Acc.  &  Best Rank &  Avg. Rank \\
       \hline
          Boltzmann  & [12220111] & 92.39 & $92.30 \pm 0.10$ & 406 & 453 \\
          Independent & [22120211] & 92.44 & $92.29 \pm 0.21 $ & 347 & 412 \\
           MCTS  & [22221210] & 92.74 & $92.51 \pm 0.18 $ & 80 & 246 \\
         Uniform & [21222220]  & 92.79 & $92.58 \pm 0.20$ & 56 & 197 \\
         MCTS + Reg. \cite{su2021prioritized} & [12222222] & 92.92  & $92.67 \pm 0.18 $ & 21 & 112 \\
          MCTS + Learned (ours) &
         [22212220] & 93.13 & $92.97 \pm 0.12$ & 1 & 6 \\
           
         \hline
         Best &  [22212220] & 93.13 & - & 1 & -  \\ 
    \end{tabular}
\end{table*}

\begin{table}
\label{resut:imagenet}
    \centering
    \caption{\textbf{Comparison of accuracy and computational cost on ImageNet classification task.}  The architecture are searched on MobilenetV2-based search space. We consider light weight models with target budget of around 280 MFLOPs. In the top part of the table we report results of other NAS methods, while at the bottom we report results of our baselines and our approach.
    }    
    \vskip 0.1in
    \label{table:config}
    \begin{tabular}{l c c c c}
 
        Method              & \begin{tabular}{@{}c@{}}Best \\ Acc. \end{tabular} &   
        \begin{tabular}{@{}c@{}}FLOPs \\ (M) \end{tabular} & \begin{tabular}{@{}c@{}}Params. \\ (M) \end{tabular} & \begin{tabular}{@{}c@{}}GPU \\ days \end{tabular} \\  
        \hline 
        MobileNetV2 
        & 72.0 & 
        300 & 3.4 & - \\
        MnasNet-A1
        & 75.2 & 
        312 & 3.9 & 288 \\
        SCARLET-C
        & 75.6 & 
        280 & 6.0 & 10 \\
        GreedyNAS-C
        & 76.2 & 
        284 & 4.7 & 7 \\
        MTC\_NAS-C
        & 76.3 & 
        280 & 4.9 & 12 \\      
        \hline
        Uniform  & $72.2  $ & 
        277 & 4.6 & $ \sim 5  $\\
        Boltzmann  & $73.1  $ & 
        278 & 4.7 & $ \sim 5  $\\
        MCTS + Reg. 
        & $76.0  $ & 
        280 & 4.9 & $ > 12  $\\

        Ours          & $76.7  $  & 
        280 & 4.9 &  $ \sim 7  $\\

    \end{tabular}
\end{table}

\textbf{Branching quality and NAS convergence.}
To show that good branching can speed up NAS, we compare NAS with our learned tree with default tree and a binary tree created from a random matrix (see Fig. \ref{supernet-iter}(right)). While the average accuracy of supernet increases in general over epochs, the branching quality can affect how the search space is explored.
After a warm-up period for UCT, our tree consistently outperforms default and random tree. This suggests that the quality of branching is important in learning how to explore the search space more efficiently and a low quality hierarchy can lead to poor performance.


\textbf{Alternative branching.}
While using the output matrix to design the branching is well-performing, it requires some initial training of the supernet. We explore using two types of encodings as a zero-cost proxy to calculate the similarity matrix. Representing an architecture as a graph, the general encoding is the adjacency matrix, corresponding to the edges (or one-hot encoding of operations per layer). 
We also consider the categorical representation of the one-hot encoding as vectors as an alternative.

As shown in Table \ref{Toy-alt-branch}, these two encoding techniques do not perform as well as our proposed approach. However, their performance is better than the most common techniques shown in Table \ref{Toy-comparison}.
We note that naively calculating the distances (same weight for all layers) is equivalent to considering each layer as an independent variable, while in fact the posterior layers have less importance than early layers. Therefore, we also considered weighting each layer $l$ exponentially with $1/2^l$, when calculating the similarity matrix, which leads to slight increases in performance. We hope in future work it will be possible to learn a good tree structure without pre-training the supernet.


\subsection{NAS-Bench-Macro Search Space}
This benchmark based on MobileNetV2 \cite{sandler2018mobilenetv2} blocks consist of 8 layers and operation set \{\textit{Identity}, \textit{MB3\_K3}, \textit{MB6\_K5\}} resulting in $3^8=6,561$ (3,969 unique) architectures. 
Table \ref{Bench-Macro-comparison} presents results of several sampling based NAS methods. 
Our method yields the best architecture of the search space. 
For each method, we use the best reward and present an ablation on different rewards in Appendix \ref{app:MCTS_reward}.

\subsection{Search on ImageNet}
ImageNet \cite{ILSVRC15} consists of 1.28 million training images in 1000 categories. In our experiments, we use 50k images of validation set as the test data to compare with other methods. To accelerate our training we use mixed precision and FFCV \cite{leclerc2023ffcv} library. We use similar macro search space to \cite{su2021prioritized,you2020greedynas,chu2021fairnas,guo2020single}, based on MobileNetV2 \cite{sandler2018mobilenetv2} blocks with optional Squeeze-Excitation (SE) \cite{hu2018squeeze} module. The total operation choice per layer is 13 resulting in $13^{21}$ search space size for 21 layers. The choices are convolution kernel size of $\{3, 5, 7\}$ and expansion ratio of $\{3, 6\}$, identity and SE option. 

Similar to \cite{su2021prioritized}, we define a FLOPs budget for our search. We leverage the fact that FLOPs can be used as a zero-cost proxy for architecture performance \cite{chen2021bench} and search only within a certain range of target budget ($[0.99,1] \times$ budget) by sampling architectures and discard those not within the budget. To compare directly with \cite{su2021prioritized}, we set a budget of $280$ MFLOPs. In Table \ref{resut:imagenet}, we compare our method with several NAS approaches (taken from \cite{su2021prioritized}) on the upper part of the table, while we compare with our sampling based approaches on the bottom part. Note that MCTS + Reg. is our re-implementation of \cite{su2021prioritized}, with some minor performance differences.
Our method yields an architecture that provides high accuracy with a limited GPU cost. We attribute this advantage to the learned structure of the tree, that allows a quicker learning of the promising architectures.

\section{Conclusion}
\label{con-limit}
In this work, we introduce a novel method to design a hierarchical search space for NAS. we highlight the shortcomings of node independence assumption used in popular NAS methods and the impact of hierarchical search space design on search quality and efficiency. We show that by simply learning the proper hierarchy, we can achieve state of the art results with MCTS without requiring further regularization and established our method empirically the by extensive evaluation on CIFAR10 and ImageNet. 
\vspace{-10pt}

\section*{Acknowledgments}
This work was supported by the Natural Sciences and Engineering Research Council of Canada and the Fonds de recherche du Québec – Nature et technologies. We also thank the Digital Research Alliance of Canada (alliancecan.ca) for the use of their computing resources.

\bibliographystyle{unsrt}  
\bibliography{references}  






\newpage
\appendix
\onecolumn
\section{More Related Work}
\label{app:related}
\paragraph{Search space design} Common NAS search spaces can be categorized as Micro (cell-based), Macro, and mulit-level. 
Micro search spaces \cite{zoph2018learning,liu2018darts} focus on yielding the optimal architecture by finding the operations that can produce the best model  inside a cell or block. The cell is then stacked to form the entire network, while the outer skeleton of the network is often controlled manually by including reduction cells. This was inspired by observing that the state-of-the-art manual architectures were formed by repetition of a certain structure and helped to reduce the complexity of search space to a manageable level \cite{white2023neural}. 

Therefore, the objective of this approach is to find a cell that works well on all parts of the network, which might be suboptimal. This neglects to explore non-homogeneous architectures and diminishes the capability of NAS to find novel architectures.
Macro search space \cite{su2021prioritized} instead searches 
for the outer skeleton of the network while fixing the operations at micro level. This can include architecture parameters such as: type of layers, number of layers, or channels in layers, pooling positions etc. Finally, a mulit-level search space searches at two levels: cell and macro structure for a CNN \cite{liu2019auto} or convolution and self-attention layers for vision transformers \cite{chen2021glit}. In this work, we focus on macro search spaces as it is generally more expressive and challenging than micro search spaces.

\paragraph{NAS benchmarks}
NAS benchmarks are used to evaluate for a given method the quality and the amount of computation required to yield a solution. They have played a crucial role in the NAS community as they provide an evaluation of all architectures in a brute-force way to find the optimal solution and eliminate the need to run independently this expensive process.
The development of NAS benchmarks has also improved the reproducibility and efficiency of NAS research. Tabular benchmarks \cite{ying2019bench,su2021prioritized} are constructed by exhaustively training and evaluating (metrics such as accuracy, FLOPS, number of parameters, etc.) all possible architectures. On the other hand, surrogate benchmarks \cite{siems2020bench} estimate the architecture performance using a model which is trained on data from several trained architectures. Benchmarks have been developed for both micro \cite{ying2019bench,dong2020bench,siems2020bench} (with addition of channels \cite{dong2021nats}) and macro \cite{su2021prioritized,roshtkhari2023balanced} search spaces. 
For more details on NAS benchmarks see the survey \cite{chitty2023neural}.

\paragraph{Architecture encoding} 
Some works have shown that architecture encoding can affect the performance of NAS \cite{white2020study, ying2019bench} and good encoding of architectures enables efficient calculation of relationships or distances among architectures. The most common encoding represents the architecture as a directed acyclic graph (DAG) and adjacency matrix along with a list of operations \cite{ying2019bench, zoph2016neural}. For using performance predictors, BANANAS \cite{white2021bananas} proposes a path-based encoding instead of adjacency matrix and GATES \cite{ning2020generic} proposed a graph based encoding scheme that better mode the flow information in the network.
Encoding can also be learned by unsupervised training prior to NAS often utilizing an autoencoder \cite{li2020neural,lukasik2021smooth,lukasik2022learning,yan2020does,zhang2019d} or a transformer \cite{yan2021cate}.
In our work, we make use and compare different ways of encoding architectures for our approach as ablation and show that measuring distances based on the network output seems to be the fundamental for good results.

\section{Experimental Setup and Details}
\label{app:exp_details}

\paragraph{Sampling method details}
A summary of various methods used in our experiments (Tables \ref{Toy-comparison} and \ref{Bench-Macro-comparison}) is presented in Table \ref{table:config}.

\begin{table}
    \caption{\textbf{ Summary of sampling methods used in our experiments.}}
    \label{table:config}
    \vskip 0.1in
    \centering
    \begin{tabular}{l c c c}
       Method & Search Space Structure & Sampling Method \\
       \hline
        Uniform & Flat & Uniform (sec. \ref{method:uniform})  \\
        Independent & Flat & Nodes sampled independently (sec. \ref{method-independent}) \\
        Boltzmann & Flat & Joint prob. (sec. \ref{method-Boltzmann})  \\ 
        Mixture & Flat (partitioned) & Uniform \cite{roshtkhari2023balanced} \\ 
        MCTS & Hierarchical (def. tree) & Conditional prob. ( sec. \ref{sec-method}, Tree Search)  \\ 
        MCTS + Reg. & Hierarchical (def. tree) & Conditional prob. + regularization \cite{su2021prioritized}\\ 
        MCTS + Learned & Hierarchical (learned tree) & Conditional prob. (sec. \ref{method-our}) \\ 
    \end{tabular}

\end{table}

\paragraph{Training Algorithm} The training pipeline for our method is shown in algorithm \ref{alg:method}. First, a pre-training with random sampling of the architectures is performed in order to train an initial model $f$ with parameters $w_p$. With this model we build a pairwise matrix $D_{i,j}$ that measures the distance of configuration $i$ and $j$ on the output space of the model. With this matrix, we use agglomerative clustering to build a binary tree that represents the hierarchy that will be used for the subsequent MCTS training. During training, an architecture is sampled from the tree, where at each node Boltzmann sampling with the learned probabilities is used. Then, this architecture is used to update the model on a mini-batch of training data (for simplicity we did not include momentum in the gradient updates) and to estimate its accuracy on a validation mini-batch. The validation accuracy is smoothed with an exponential moving average and used as reward with UCT regularization for updating the node probabilities of the sampled architecture.  
Finally, after the MCTS, the best architectures are sampled from the tree by sampling the tree with $\lambda=0$.

\textbf{Tree design.}
 Consider the toy search space shown in Fig.\ref{fig:tree_design}(a), with 3 binary operations $(O_a,O_b,O_c)$ which leads to 8 distinct architectures.
  The default tree design (Fig. \ref{fig:tree_design}~(b)) as presented in sec.\ref{sec-method}, branches off the tree on operation choices per layer.

\begin{algorithm}[H]
\caption{Simplified pseudo-code of our training pipeline.}\label{alg:method}
\SetKwInOut{Input}{Input}
\SetKwInOut{Output}{Output}
\Input{$\mathcal{S}$: Search Space; $\mathcal{X}_{t},\mathcal{X}_{v}$: mini-batches of training and validation data; $f_a$: model with architecture $a$; $w_p,w$: weights of the pre-trained and final model initialized randomly; ${e}_{pt},{e}_{MCTS}$: pre-training and MCTS iterations; $\alpha$: learning rate; 
$\beta$: smoothing factor; $\lambda$ : exploration parameter of UCT. 
}

\#$pre$-$training$ \\
\While {$epochs \leq {e}_{pt}$
}{
$a \leftarrow$ sample from  $\mathcal{U}(|\mathcal{S}|)$ \\
$w_p \leftarrow$ $ w_p - \alpha \nabla_{w_p} \mathcal{L}(f_a(\mathcal{X}_t,w_p))$ \\
}
\#$build$ $the$ $search$ $tree$ \\
\For{$a^i \in \mathcal{S}$
}{
Output vector $o^i \leftarrow f_{a^i}(\mathcal{X}_v,w_p)$
}
Distance matrix $D_{ij} = dist(o^i,o^j)$ \\
Binary tree $\mathcal{T} \leftarrow aggl\_clustering(D)$ \\
\#$main$ $training$ $with$ $MCTS$ \\
\While{$epochs \leq {e}_{MCTS}$
}{

\#$sample$ $an$ $architecture$ $a$ \\
$a$ = [] \\
$node$ = "root" \\
\textbf{push}($a$,$node$) \\
\While { $not(is\_leaf(node))$ 
}{
$node \leftarrow sample(next(node))$ with Boltzmann sampling as in Eq.(\ref{eq:uct}) \\ 
\textbf{push}($a$,$node$)\\
}   
\#$update$ $model$ $w$ $and$ $accuracy$  \\
$w \leftarrow w- \alpha \nabla_w \mathcal{L} (f_a(\mathcal{X}_t,w))$\\

$accuracy \leftarrow$ Acc($f_a(\mathcal{X}_{val},w))$ 

$node \leftarrow$ \textbf{pop}($a$) \\

\#$update$ $rewards$\\
\While { $not(is\_root(node))$ 
}{
$parent \leftarrow$ \textbf{pop($a$)}\\
$C(node) \leftarrow \beta ~ C(node) + (1-\beta) ~ {accuracy}  $ \\
$R(node) \leftarrow C(node) + \lambda \sqrt{{log(|parent|)}/{|node|}} $ \\
$node \leftarrow parent$
}
}
\Output{Best architecture from $\mathcal{T}$ by sampling with $\lambda=0$}

\end{algorithm}

\subsection{Dataset and Hyperparameters}
\label{app:exp_det}
For experiments performed on CIFAR10 \cite{kocsis2006bandit} dataset, we split the training set 50/50 for NAS training and validation. To tune hyperparameters, we either performed grid search or when comparing with other works used similar hyperparameters. We used SGD with weight decay and cosine annealing learning rate schedule. Furthermore, for MCTS methods we split training iterations to roughly 40/25/35 fractions for uniform sampling/MCTS warm-up/MCTS sampling respectively. In all experiments we use $\beta=0.95$ and $\lambda = 0.5$.

\paragraph{Pooling search space} This search space introduced by \cite{roshtkhari2023balanced} is based on Resnet20 \cite{he2016deep} architecture. The only CNN parameters to search is where to perform pooling. To calculate distance matrix we trained the supernet for 300 epoch using uniform sampling with batch size 512, learning rate 0.1 and weight decay 1e-3. For search we trained for 400 epochs with batch size 256, learning rate learning rate 0.05 and temperature $T$ is set to linearly annealing schedule (0.02, 0.0025). Since this search space is small we only consider nodes with max probabilities and report the it as the final architecture.

\paragraph{NAS-Bench-Macro search space} This benchmark introduced by \cite{su2021prioritized} is based on  MobileNetV2 \cite{sandler2018mobilenetv2} blocks. The supernet pre-training is performed for 80 epochs with batch size 512 and learning rate 0.05. For search we use batch size of 256 for 120 epochs with and $T$ linearly annealing from 0.01. At the ends of training, we sample 50 architectures from the tree and report the best as final architecture.

\paragraph{ImageNet}
To accelerate our training in ImageNet experiments, we use mixed precision and FFCV \cite{leclerc2023ffcv} library. We sample architectures within a FLOPs budget and discard those outside of it. We train for 100 epochs with SGD and cosine annealing learning rate. Other training strategies are similar to experiments on CIFAR10.

\section{Additional Results}

\subsection{Reward for MCTS}
\label{app:MCTS_reward}
The most common rewards used for NAS algorithms is accuracy and loss. While loss is differentiable, accuracy is more aligned with the objective of NAS. Furthermore, either the training or validation can be used to calculate the reward. Instead of absolute values, a relative training loss metric was used in \cite{su2021prioritized} to account for unfair reward comparison at different iteration of supernet training. In Table \ref{Bench-Macro-reward} for NAS-Bench-Macro, we explore some common combination of options to estimate the reward. 
In all setting our approach performs on par or slightly better than Default Tree + Regularization. For both methods it seems that using the accuracy on the validation set as metric is the best. However, while for our approach the best performing configuration is obtained with the absolute metric, for \cite{su2021prioritized} the relative metric seems slightly better.   



\begin{table}[tb]
    \caption{\textbf{CIFAR10 results on NAS-Bench-Macro~\cite{su2021prioritized}} search space with several various rewards. Relative rewards are calculated according to \cite{su2021prioritized}. The rewards can be can be calculated on either training or validation set.}
    \label{Bench-Macro-reward}
    \vskip 0.1in
    \centering
    \begin{tabular}{l c c c c c c c c}
       \begin{tabular}{@{}c@{}}Search \\ Structure \end{tabular} & Metric & \begin{tabular}{@{}c@{}} Reward \\ Data \end{tabular} & \begin{tabular}{@{}c@{}}Reward \\ Measure \end{tabular} & Arch. &  \begin{tabular}{@{}c@{}}Best \\ Acc. \end{tabular}  &  \begin{tabular}{@{}c@{}}Best \\ Rank\end{tabular} &  \begin{tabular}{@{}c@{}}Avg. \\ Rank\end{tabular} \\
       \hline
            Default Tree + Reg & rel.  & train & loss  & [22121222] & 92.74 & 85 & 97 \\
           Learned Tree (ours) & rel. & train & loss & [22122220] & 92.78 & 61 & 67  \\
             \hline
             Default Tree + Reg  & abs. & train & acc.  & [22121210] & 92.55 & 227 & 278   \\
            Learned Tree (ours) & abs. & train & acc. & [22110222] & 92.56 & 209 &  301 \\
            \hline
           Default Tree + Reg  & rel. & val. & loss  & [22222022] & 92.71 & 98 &  120 \\
           Learned Tree (ours) & rel. & val. & loss & [21211220] & 92.76 & 71 & 95  \\
                      \hline
          Default Tree + Reg  & rel. & val. & acc.  & [12222222] & 92.92 & 21 &  112 \\
            Learned Tree (ours) & rel. & val. & acc. & [22212200] & 92.94 & 19 &  67 \\
           \hline
          Default Tree + Reg  & abs. & val. & acc.  & [22221200] & 92.86 & 34 & 54  \\
            Learned Tree (ours) & abs. & val. & acc. & [22212220] & 93.13 & 1 & 6  \\
    \end{tabular}
\end{table}

\subsection{Distance Matrices}
\begin{figure}
  \centering
  \includegraphics[width=0.3\textwidth] {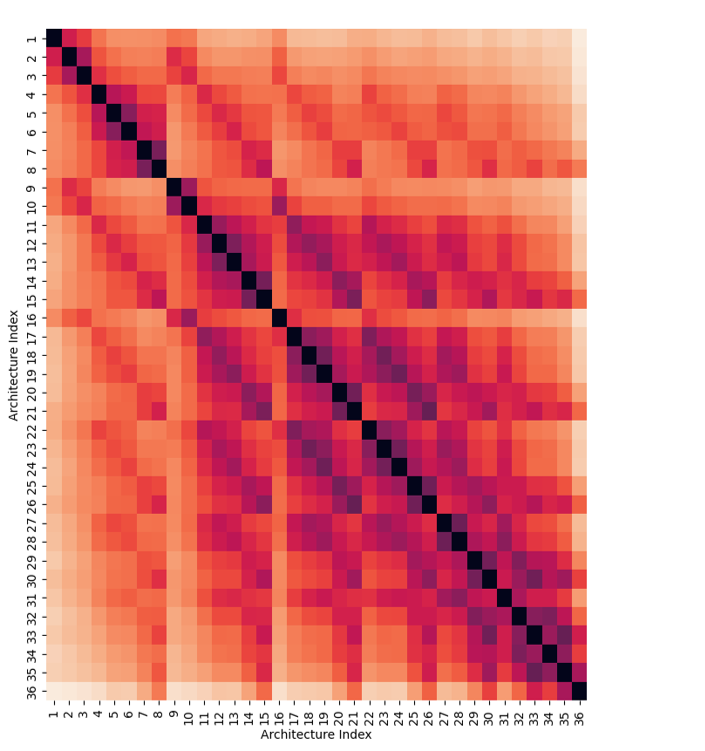}
  \includegraphics[width=0.3\textwidth]
  {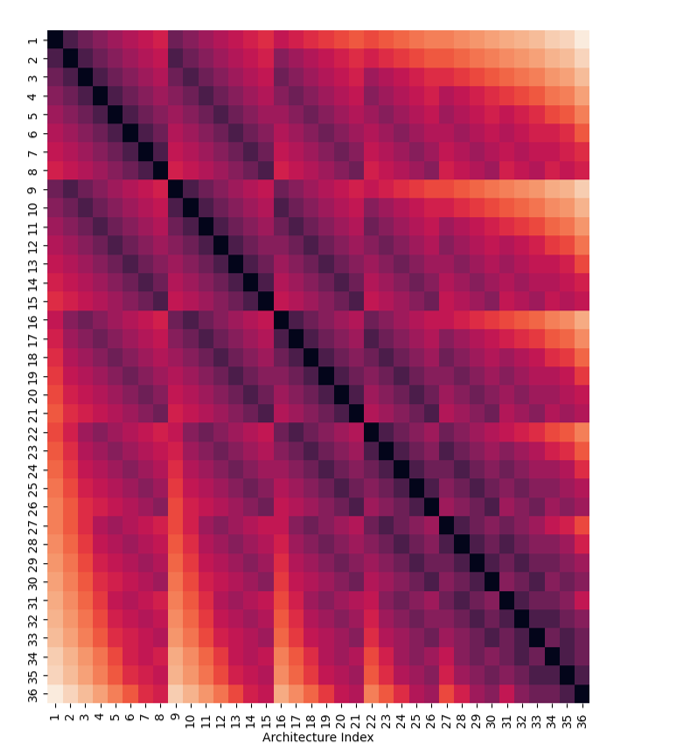}
  \includegraphics[width=0.3\textwidth]
  {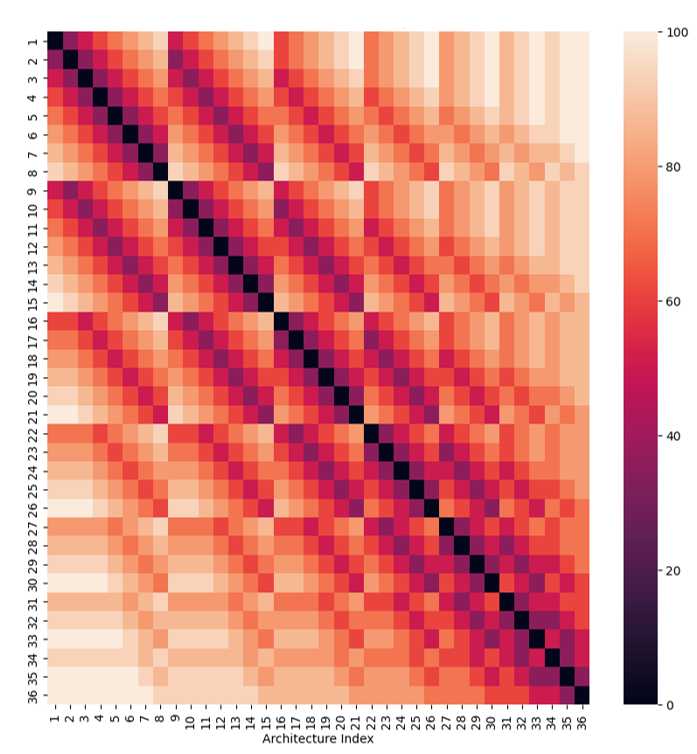}
  \caption{ \textbf{Normalized distance matrices calculated with various methods.} (left) Distance matrix calculated from output vectors (our method) ; (middle) From vector encoding ; (right) From one-hot encoding. The architecture indices on leaves correspond to indices used in Pooling benchmark \cite{roshtkhari2023balanced}.
  }
  
  \label{fig:distance_mat}
  
\end{figure}

We visualize the distance matrices calculated using output vector and various encoding in Fig. \ref{fig:distance_mat}.

\subsection{Tree Visualizations}
\begin{figure}
  \centering
  \includegraphics[width=0.3\textwidth] {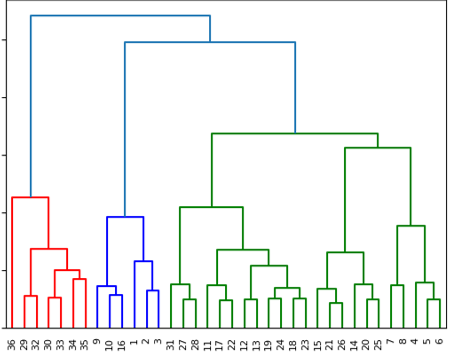}
  \includegraphics[width=0.3\textwidth]
  {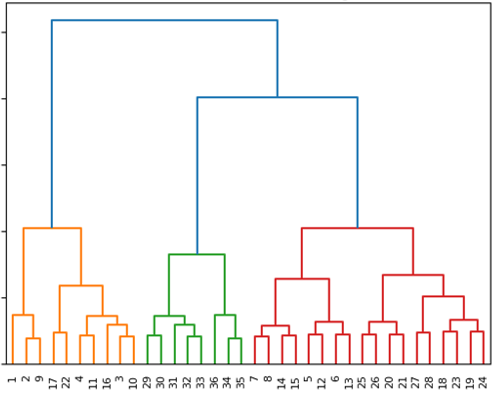}
  \includegraphics[width=0.3\textwidth]
  {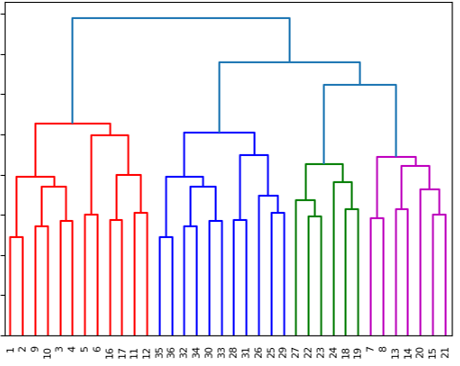}
  \caption{ \textbf{Tree branching for Pooling search space by hierarchical clustering.}  The architecture indices on leaves correspond to indices used in Pooling benchmark \cite{roshtkhari2023balanced} (left) Tree learned from output vectors (our method) ; (middle) From vector encoding ; (right) From one-hot encoding.
  }
  
  \label{fig:tree}

\end{figure}

For pooling search space, we visualize tree structure based on our proposed method and architecture encodings in Fig. \ref{fig:tree}. The tree is presented with architecture indices on leaves. The architecture indices and the  corresponding performance can be found in \cite{roshtkhari2023balanced}.


\end{document}